# Using Summarization to Discover Argument Facets in Online Idealogical Dialog

Amita Misra, Pranav Anand, Jean Fox Tree, and Marilyn Walker
UC Santa Cruz
Natural Language and Dialogue Systems Lab
1156 N. High. SOE-3
Santa Cruz, California, 95064, USA
amisra2|panand|foxtree|mawalker@ucsc.edu

## Abstract

More and more of the information available on the web is dialogic, and a significant portion of it takes place in online forum conversations about current social and political topics. We aim to develop tools to summarize what these conversations are about. What are the CENTRAL PROPOSITIONS associated with different stances on an issue; what are the abstract objects under discussion that are central to a speaker's argument? How can we recognize that two CENTRAL PROPOSITIONS realize the same FACET of the argument? We hypothesize that the CENTRAL PROPOSITIONS are exactly those arguments that **people** find most salient, and use human summarization as a probe for discovering them. We describe our corpus of human summaries of opinionated dialogs, then show how we can identify similar repeated arguments, and group them into FACETS across many discussions of a topic. We define a new task, ARGUMENT FACET SIMILARITY (AFS), and show that we can predict AFS with a .54 correlation score, versus an ngram system baseline of .39 and a semantic textual similarity system baseline of .45.

## 1 Introduction

In the wake of the Penn TreeBank, much progress has been achieved in processing the monologic, informational language characteristic of newswire text. But an increasing share of the text data on the web is unlike newswire in a variety of ways: it is dialogic, opinionated, argumentative. And while some of these dialogs may be a little more than flame wars, a significant portion involve contentful, rea-soned disputes on important social and political topics, as exemplified by the forum snippets in Figs. 1 and 3. Studying data like this will undoubtedly help us to understand dialogic and informal argumentative language in general. And, indeed, previous work (Abbott et al., 2011; Somasundaran and Wiebe, 2010) has examined the structure of these discussions – e.g., the argumentative discourse relation a post bears to its parent (agreeing or disagreeing), or the stance that a person takes on an issue.

| PostID:Turn |
|---|
| **S1:1** Agreed She is ignoring my religious freedom and trying to institute her religion into law. **The law that will bar my family from legal protections. It won't protect her marriage but will bar me and my people from from being full citizens**. She isn't protecting marriage but perserving her heterosexual privledge. |
| **S2:1** How on earth is she impeding on you religious freedom? She isn't trying to take away your right to any religious ceremony. With such a wide-open standard of what constitutes religious freedom that you seem to have, any legislation could be construed as imposing on religious freedom. |
| **S1:2** Because it is her religious belief that marriage is between a man and a woman. **My religious belief is that marriage is between two people that love each other regardless of sex**. She is tying to place her religious belif into law over mine. Who gets hurt here? If my religious belief is put into law she can still marry the person of her choice. If her religious belief gets put into law she can still marry the person of her choice but I do not get to. So I and my people are hurt by codifing her religious belief into law. She is trying to keep gay people out of marriage and thus preserve her heterosexual privledge. |
| **S2:2** But by that definition, either one could be viewed as impeding on religious freedom, including your view impeding on hers ! We don't define imposing on religious freedom on the basis of having different ideals. **It doesn't effect your religion or religious freedom if you don't get benefits under gay marriages**. You can argue in other ways, on other basis, but the idea that not giving gays marriage benefits is imposing on religious freedom is an empty " argument ". |

Figure 1: Gay Marriage Dialog-1.

Our goal here is to develop techniques to recognize the specific arguments and counterarguments people tend to advance, and group them across discussions into the FACETS on which that issue is ar-

430

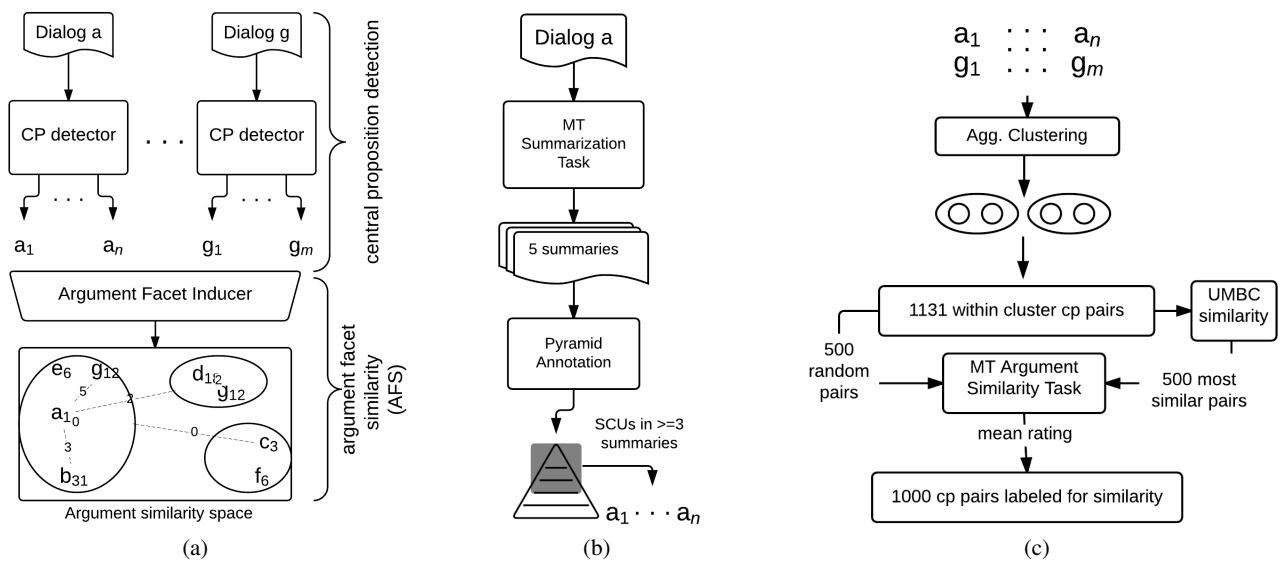

Figure 2: The overall engineering architecture of our approach. (a) Basic engineering approach for extracting CENTRAL PROPOSITIONS and clustering them into argument FACETS across several dialogs; (b) Workflow for 'detecting' central propositions via pyramid evaluation of multiple summaries; (c) Workflow for obtaining gold-standard labels for AFS task.

gued across the population at large. Recognizing the FACETS of an argument automatically entails at least two subtasks, as schematized in Fig. 2a.

| PostID:Turn |
|---|
| **S1:1** Certainly not yours. You should know that I am for no marriage in government. It should be left to a religious institution where it will actually mean something. The states should then go back to doing something that actually makes sense and doesn't reward people like Britney Spears for being white trash. |
| **S2:1** That is all well and good, but it is not the religious ceremony and sanction that gays are looking for. They already have that; there are churches that perform same-sex marriages. **It is the civil benefits that are at issue. Are you saying you would be in favor of foregoing ALL the legal rights and benefits you are afforded by marriage? For example: *Assumption of Spouse's Pension *Automatic Inheritance *Automatic Housing Lease Transfer *Bereavement Leave....** What do you say? |
| **S1:2** yeah I know. I'm saying that there should be a better system. For example, if you had a best friend who you are roommates with... both hetero for the sake of argument... and never wish to get married then **could they get some of the benefits you described**? |

Figure 3: Gay Marriage Dialog-2.

First, there must be a system, the CENTRAL PROPOSITION detector, that can extract the most essential arguments in a particular conversation. Example CENTRAL PROPOSITIONS in Figs. 1 and 3 are provided in bold. Second, there must be another system, the ARGUMENT FACET inducer, that relates these conversation-specific arguments to each other in terms of FACETS, e.g. that identifies the two specific central propositions in Figs. 1 and 3 about "legal protections" and "civil benefits" as the same (abstract) FACET, namely that same-sex marriage is about getting the civil rights benefits of marriage.

We first focus on the question of extracting reliable data for central propositions. See Fig. 2b. We propose that the CENTRAL PROPOSITIONS of a dialog are exactly those arguments that **people** find most salient, which is naturally reflected by their summarization behavior. We then apply the Pyramid method, by which the CENTRAL PROPOSITIONS bubble up to the highest tiers of the pyramid, thereby allowing us to identify them. With the central propositions in hand, we proceed to build the argument facet inducer. We introduce a new task of ARGUMENT FACET SIMILARITY (**AFS**). We discuss how AFS is similar to, but different than SEMANTIC TEXTUAL SIMILARITY (**STS**) (Agirre et al., 2012; Jurgens et al., 2014; Agirre et al., 2013; Beltagy et al., 2014; Han et al., 2013).

Sec. 2 provides a more detailed overview and description of our method, and the data that it produces. Sec. 3 describes our experimental setup for the **AFS** task and then presents our results. We describe a learning approach that achieves correlations of .54 on the **AFS** task, as compared to a baseline correlation of .45 using off-the-shelf modules that

431

are competitive in **STS** tasks. We delay a detailed discussion of related work to Sec. 4 when we can compare it to our own approach. Sec. 5 summarizes the paper and discusses future work.

## 2 Experimental Method

Fig. 2 summarizes our overall method for producing the summary corpus and then extracting arguments and clustering them into FACETS. Our method consists of the following steps:

**S1**: Dialog Selection.
**S2**: MT summarization of dialogs selected in S1.
**S3**: Pyramid annotation of summaries produced by S2 and selection of top-tier pyramid labels as CENTRAL PROPOSITIONS for individual dialogs.
**S4**: Clustering of CENTRAL PROPOSITIONS from S3.
**S5**: MT ARGUMENT FACET SIMILARITY task, using clusters from S4.
**S6**: Train and test a predictor for ARGUMENT FACET SIMILARITY (Sec. 3).

We explain these steps in more detail below.

**S1: Dialog Selection.** We use the publicly available Internet Argument Corpus (IAC) (Walker et al., 2012). We use the links in the meta-data to extract a sequence of turns to build two-party dialog chains like those in Figs. 1 and 3. We extracted 85 dialogs for the topic gay marriage from an original corpus of 1292 discussion threads using these criteria:

- **Number of turns per contributor**: We want dialogs in which substantive issues were discussed, so we extract dialogs with at least 3 turns per conversant that present at least 2 different perspectives on an issue.

- **Author**: Some authors post frequently and would dominate the corpus if we use random selection. To get richer, more diverse dialogs expressing different perspectives, we only select a single dialog between any particular pair of authors from a discussion thread.

- **Word Count in a post**: Some posts are long. To make it practical to collect dialog summaries, we extract dialogs where the number of words per turn is less than 250.

**S2: MT Summarization Task.** The summarization task was run on Mechanical Turk. To get good

| S1 thinks that the government should stay out of marriage and that it should be left to religious institutions. He thinks there needs to be a better system and that single people are the ones that are harmed the most by marriage laws because they are unable to get any of the benefits that married people do even if they want them, or it is important to their situation. S2 says religious ceremonies aren't what gay people want because they already can have them via churches. They want the rights and to keep the government out would be to give up those rights. If single people want those rights they should get married, but he thinks you should be free to marry who you wish. |
| --- |
| The issue here is whether government or religion should decides the principles of marriage, and who is allowed to get married.<br>Speaker one believes that leaving it up to religions groups does not satisfy what gays are looking for. They are searching for the civil benefits that come with a marriage and would like to be treated equally in that respect. The speaker believes gay should be able to marry a person of their choice and get equal rights. Speaker two opinions that there should indeed be a better system for marriage benefits and that it is all "single" people that get screwed over by marriage's current stature. Speaker two believes that gay people should marry a woman if they want the same rights. |

Figure 4: Two of the 5 Summaries for Dialog-2.

quality summaries, workers completed a qualification test involving summarizing a sample dialog. Workers were instructed to summarize according to dialog length: dialogs under 750 words in 125 words, and those above 750 in 175 words. We use 45 dialogs in this study and save the other 40 for future work. We collect 5 summaries for each dialog resulting in a dataset of 225 summaries. Fig. 4 provides 2 of the 5 summaries collected for the dialog in Fig. 3.

**S3: Pyramid Annotation.** We trained three undergraduates to annotate summaries to produce pyramids. We hypothesize that we can use the Pyramid method to induce the FACETS of a topic across a set of dialogs (Nenkova and Passonneau, 2004). The annotation of Pyramids seeks to uncover the common elements, or summary content units (SCUs), across several summaries (in our case, 5). Each SCU identifies a set of spans that are semantically equivalent. Each SCU also has a unique annotator-generated label that reflects the semantic meaning of the contributions. Because our aim here is to focus on argument propositional content, the annotators were instructed to keep only the main proposition in the SCU as the label, ignoring any attributions or other types of content. See Table 1. Once annotation is complete, the SCUs are ranked based on their frequency across all of the summaries, as shown by the Tier in Fig. 5, which includes data from the two summaries in Fig. 4.

| Contributor | S1 points to the trend to legalize gay marriage in western countries such as Netherlands, Belgium, and most of Canada |
|---|---|
| Contributor | S1 refutes this assertion, citing a number of countries which recognize same-sex marriage. |
| Contributor | He states the US is more similar to Anglo nations and in many of those gay marriage is legal. |
| Label | A number of countries recognize same-sex marriage. |

Table 1: A sample label after removing the attributions from the SCU contributors.

**S4: SCUs to clusters.** The pyramid structure directly reflects the content that the annotators deem most important in the original dialog. We are interested in the content that bubbles to the top across all the dialogs. We take the Tier 3 and above SCUs as our CENTRAL PROPOSITIONS, and extract the labels of those SCUs. This gives a total of 329 SCU labels. In what follows we treat a cluster of CENTRAL PROPOSITIONS as a FACET label, just as a synset concept in WordNet is labeled by its members.

The purpose of AFS, then, is to provide a similarity metric on these SCU labels. As described below (and sketched in Fig. 2c), we used Mechanical Turk to provide similarity scores between pairs of SCU central propositions. Although, in principle, we could have asked about all possible pairs of the 329 CENTRAL PROPOSITIONS, most pairs are likely to be unrelated, and so we used an initial clustering algorithm to help reduce the work and cost.

To group similar arguments, we performed clustering across our 329 labels. We performed Agglomerative Clustering using Scikit-learn (Agg Clustering in Fig. 2c). (Pedregosa et al., 2011). It recursively merges the pair of clusters that minimally increases a given linkage distance. We used cosine similarity as the distance measure with average linkage criteria. To focus on topic-specific cues, the clustering was performed using only nouns, verbs and adjectives. After generating all pairwise combinations within a cluster, this approach yielded 1131 argument pairs used in the Mechanical Turk AFS task. See Fig. 2c.

---

**Instructions**
We would like you to classify each of the following sets of pairs based on your perception of how SIMILAR the arguments are, on the following scale, examples follow.
(5) Completely equivalent, mean pretty much exactly the same thing, using different words.
(4) Mostly equivalent, but some unimportant details differ. One argument may be more specific than another or include a relatively unimportant extra fact.
(3) Roughly equivalent, but some important information differs or is missing. This includes cases where the argument is about the same FACET but the authors have different stances on that facet.
(2) Not equivalent, but share some details. For example, talking about the same entities but making different arguments (different facets)
(1) Not equivalent, but are on same topic
(0) On a different topic

**Facet**: A facet is a low level issue that often reoccurs in many arguments in support of the author's stance or in attacking the other author's position. There are many ways to argue for your stance on a topic. For example, in a discussion about the death penalty you may argue in favor of it by claiming that it deters crime. Alternatively, you may argue in favor of the death penalty because it gives victims of the crimes closure. On the other hand you may argue against the death penalty because some innocent people will be wrongfully executed or because it is a cruel and unusual punishment. Each of these specific points is a facet.
For two utterances to be about the same facet, it is not necessary that the authors have the same belief toward the facet. For example, one author may believe that the death penalty is a cruel and unusual punishment while the other one attacks that position. However, in order to attack that position they must be discussing the same facet.

Figure 6: Instructions for AFS MT HIT.

**S5: MT Argument Facet Similarity HIT.** Fig. 6 shows the instructions defining AFS for the MT HIT. Inspired by the scale used for STS, we collected annotations on a 6 point scale. One crucial difference in our formulation was a desire to capture similarity in FACET and argument simultaneously. The use of the value 3 for 'same FACET, contradictory stance' was a well-thought decision in the definition of AFS.

| SCU Label | Used by summarizer? | | | | | Tier |
|---|---|---|---|---|---|---|
| | 1 | 2 | 3 | 4 | 5 | |
| Gay couples are interested in the rights and benefits associated with marriage. | ✓ | ✓ | ✓ | ✓ | ✓ | 5 |
| Gay people should be able to marry a person of their choice and get equal rights. | ✓ | ✓ | ✓ | ✓ | ✓ | 5 |
| Government should not be involved in marriage and marriage should be left to religious institutions. | ✓ | ✓ | ✓ | ✓ | ✓ | 5 |
| Discussion on the civil benefits of marriage and the rights of marriage. | ✓ | | ✓ | ✓ | ✓ | 4 |
| Gay couples are unable to get any benefits that married people do. | ✓ | ✓ | | ✓ | ✓ | 4 |
| There should be a better system for marriage benefits. | | ✓ | ✓ | ✓ | ✓ | 4 |
| Religious ceremonies are not what gay people want. | | ✓ | | ✓ | ✓ | 3 |
| Single people are the ones that are harmed the most by marriage laws. | ✓ | ✓ | ✓ | | | 3 |
| Gay people should marry the opposite sex if they want the same rights. | | | ✓ | ✓ | | 2 |
| Gays have religious ceremonies already can have them via churches | | | ✓ | | | 1 |
| Relation to the issues by consideration of the case of a life-long bachelor uncle | ✓ | | | | | 1 |

Figure 5: Pyramid for Dialog-2. SCU labels in Tiers 3-5 are assumed to be the CENTRAL PROPOSITIONS.

Just as two words can only be antonyms if they are in the same semantic field, two arguments can only be contradictory if they are about the same FACET. Thus, we instruct annotators to give a score of 3 to opposing arguments on the same FACET.

The task was put on Mechanical Turk using two separate batches. For the first batch we randomly selected 500 pairs from our pairs dataset of 1131 pairs. However, our subsequent impression was that the clustering had not filtered out enough of the unrelated pairs (score 0-1). For the second batch we selected the top 500 pairs according to the UMBC similarity score (Han et al., 2013). This gave us a final pair dataset of 1000 pairs. Since AFS is a novel and subjective task, workers took a qualification test. Then each pair was annotated by 5 workers, and one of the authors provided gold standard labels. The HIT allowed 5 AFS judgements per hit, thus the number of pairs annotated by a worker varies from 5 to 1000.

To increase reliability, we removed the annotations from those workers who had attempted less than 4 hits (20 pairs) and had the lowest pairwise correlations with our gold standard annotation. Our final AFS score was the average score across all the annotators. The final AFS score correlated at .7 with our gold standard annotation, showing that the AFS similarity task is well-defined, and understandable by minimally trained annotators on MT. Table 4 provides typical examples of argument pairs and their MT AFS score, along with the predicted scores from some of our models. We discuss the AFS values and interesting cases in Sec. 3 below.

## 3 Machine Learning Experiments and Results

Given the data collected above, we defined a supervised machine learning experiment with AFS as our dependent variable and different collections of features inspired from previous work as our independent variables.

### 3.1 Features

**NGRAM overlap.** This is our primary baseline. For each argument, we extracted all the unigrams, bigrams and trigrams, and then counted how many were in overlap across the two arguments. For unigrams we did not include stop words. Stemmed Ngrams were used to get better overlap.

**UMBC.** This is our secondary baseline. This feature is the Semantic Textual Similarity obtained using UMBC Semantic Similarity tool (Han et al., 2013)

**DISCO Distributionally Similar Category.** We used the distributional similarity tool DISCO with the pre-computed English Wikipedia word space (Kolb, 2008). We extract the top 5 distributionally similar nouns, verbs, and adjectives for each argument. For each argument pair, three vector pairs (over nouns, verbs, and adjectives) are created with this extended vocabulary. Stemming was performed and cosine similarity between these vector pairs was calculated.

**LIWC Category.** This feature set is based on the Linguistics Inquiry Word Count tool (Pennebaker et al., 2001). To tune these features, we first used a set of gay marriage posts from websites such as CreateDebate and ConvinceMe to extract relevant LIWC categories. We supplemented this data with gay marriage posts from 4forums, but excluded the discussion threads in our dialog corpus. From this data, we extracted the LIWC categories most frequent nouns, verbs and adjectives. For the verbs category, we excluded the verbs present in the NLTK stop word list. We retained only semantically rich categories such as Biological Processes, Causation, Cognitive Processes, Humans, Negative Emotion, Positive Emotion, Religion, Sexual, and Social Processes. The score for this set was the LIWC category overlap count across pairs for each category.

**ROUGE Scores.** ROUGE is a family of metrics to determine the quality of a summary by comparing it to other ideal summaries (Lin, 2004). It is based on a number of overlapping units such as n-gram, word sequences, and word pairs. This feature includes all of the rouge f-scores available via the package at https://pypi.python.org/pypi/pyrouge/0.1.0.

### 3.2 Results

Our aim is to predict the similarity among repeated arguments across many discussions in online social and political debate forums, a task we have dubbed ARGUMENT FACET SIMILARITY (**AFS**). Given the CENTRAL PROPOSITIONS from the CP detector (see Fig. 2a), we need to train an argument FACET inducer. We define AFS as a regression problem and evaluate support vector regression and linear regression for 10-fold cross validation using the Weka machine learning toolkit (Hall et al., 2005).

| Classifier | RMS | MAE | R |
|---|---|---|---|
| SMO | 1.0208 | 0.8019 | 0.532 |
| Linear Regression | 0.9996 | 0.8003 | 0.540 |

Table 2: Support Vector and Linear Regression. RMS: Root Mean Squared Error, MAE: Mean Absolute Error, R: Correlation Coefficient.

Table 2 shows that the results for support vector regression are worse than the linear regression model using our proposed features combined with UMBC, hence we focus hereon on linear regression. Table 3 provides the correlations, MAE, and RMS values for models produced using various sets of features. We considered two baselines, simple Ngram overlap and the off-the-shelf UMBC STS metric (Han et al., 2013). In general, we found that Ngram overlap (Row 1) performed best alone of our features, but falls short of the UMBC baseline (Row 2). It is interesting that Ngram alone outperforms distributional measures (which Conrad & Wiebe found most helpful) as well as Rouge (which contains metrics insensitive to linear adjacency).

| Row | Feature Set | R | MAE | RMS |
|---|---|---|---|---|
| 1 | **NGRAM (N)** | 0.39 | 0.90 | 1.09 |
| 2 | **UMBC (U)** | 0.46 | 0.86 | 1.06 |
| 3 | **LIWC (L)** | 0.32 | 0.92 | 1.13 |
| 4 | **DISCO (D)** | 0.33 | 0.93 | 1.12 |
| 5 | **ROUGE (R)** | 0.34 | 0.91 | 1.12 |
| 6 | **N-U** | 0.47 | 0.85 | 1.05 |
| 7 | **N-L** | 0.45 | 0.86 | 1.06 |
| 8 | **N-R** | 0.42 | 0.88 | 1.08 |
| 9 | **N-D** | 0.41 | 0.89 | 1.08 |
| 10 | **U-R** | 0.48 | 0.84 | 1.04 |
| 11 | **U-L** | 0.51 | 0.83 | 1.02 |
| 12 | **U-D** | 0.45 | 0.86 | 1.06 |
| 13 | **N-L-R** | 0.48 | 0.84 | 1.04 |
| 14 | **U-L-R** | 0.53 | 0.81 | 1.00 |
| 15 | **N-L-R-D** | 0.50 | 0.83 | 1.03 |
| 16 | **N-L-R-U** | 0.54 | 0.80 | 1.00 |
| 17 | **N-L-R-D-U** | 0.54 | 0.80 | 1.00 |

Table 3: Results for Different Individual Features and Feature Combinations.

Table 3, Row 15, shows that the best correlation that is achievable without UMBC is the combination of Ngram, LIWC, ROUGE and DISCO (NLRD). This combination significantly improves over the UMBC baseline of 0.46 to 0.50 (paired *t*-test, $p < .05$).

We then tested combinations of of features to determine which feature sets are complementary. LIWC + NGRAM is significantly different than NGRAM alone ($p < 0.01$), and ROUGE + NGRAM is significantly different than NGRAM alone ($p = 0.03$), but DISCO does not add anything ($p = 0.2$). This shows that LIWC and ROUGE features complement Ngram features. Other combinations of interest are NGRAM + LIWC (Row 7) which amazingly performs as well as UMBC while UMBC includes sentence alignment, a model of negation, and distributional measures (Han et al., 2013). This suggests that AFS is clearly a different task that STS. Additionally we also combined our proposed set of features with UMBC. A comparison of Row 15 (our feature set) with Rows 16 and 17 of Table 3 where we combine our features with UMBC shows that this improves the correlation further, from the UMBC baseline of 0.46 to 0.54 ($p < 0.01$.)

| Row | N | L | U | NLRD | NLRDU | MT AFS | Arg1 | Arg2 |
|---|---|---|---|---|---|---|---|---|
| 1 | 1.38 | 1.50 | **0.37** | 1.31 | 0.40 | 0.00 | everyone has the freedom of speech | service in the military |
| 2 | 2.00 | 2.02 | **1.55** | 2.33 | 1.86 | 1.14 | **gay people should be able to marry a person of their choice and get equal rights** | **referring to namecalling and violence from the original post that was opposing gay rights** |
| 3 | 2.00 | 1.29 | 2.52 | **1.37** | 1.54 | 1.33 | Constitutional right to be opposed to gay marriage as well as gay people themselves | arguing about marriage benefits between single people and married |
| 4 | 2.00 | 1.70 | 2.74 | **1.77** | 1.98 | 1.80 | people should not pick and choose what they want equal rights on. | people did not want gay marriage |
| 5 | 1.38 | 1.92 | 0.88 | **1.94** | 1.64 | 2.50 | the Republicans creating another Holocaust | No republican in leadership would call for the extermination of gays |
| 6 | 1.69 | 2.02 | **2.58** | 1.89 | 2.49 | 2.60 | **homosexuals have all the same rights as heterosexuals** | **Opposition to equal rights for gay couples.** |
| 7 | 1.83 | 2.40 | 1.46 | **2.81** | 2.51 | 3.00 | There was prejudice against gays in 1909 just as there is now | it is prejudice as opposed to religious or moral beliefs which fuel the anti-gay agenda; |
| 8 | 2.00 | 1.70 | **3.16** | 1.73 | 2.41 | 3.40 | **homosexual relationships should not compare to heterosexual marriages because only heterosexuals are legally allowed to marry** | **marriage should be between a heterosexual couple** |
| 9 | 2.00 | 2.70 | 2.09 | 2.83 | **3.03** | 3.50 | it is prejudice as opposed to religious or moral beliefs which fuel the anti-gay agenda; | when people claim religion in doing prejudice they are actually abandoning their morals |
| 10 | **2.94** | 2.02 | 2.93 | 2.18 | 2.70 | 3.50 | **gay people should be able to marry a person of their choice and get equal rights.** | **Gay couples are unable to get any benefits that married people do.** |
| 11 | 2.14 | 1.50 | **2.91** | 2.08 | 2.62 | 3.60 | **Paul Cameron is the voice of the Republicans** | **Conversation about Paul Cameron** |
| 12 | 2.63 | 3.63 | 2.60 | **3.75** | 3.57 | 4.17 | in opening this opportunity for gay marriage, the definition of marriage will change | opponents of homosexual marriage tend to argue that a change to marriage law would make it too open ended |
| 13 | **4.23** | 2.72 | 2.26 | 4.82 | 4.12 | 4.50 | AIDs was initially spread in the United States primarily by homosexuals. | No one argues the point that AIDs was spread in the United States by homosexuals. |

Table 4: Predicted Scores for each model and the Mechanical Turk AFS gold standard for selected argument pairs from the pairs dataset. Best performing model for each pair is shown in **bold**. The table is sorted by the AFS score (gold standard). The argument pairs shown in **bold** are cases where UMBC by itself beats our proposed model. KEY: Feature sets model. N = NGRAM, U = UMBC STS tool, L = Linguistic Inquiry and Word Count; R = Rouge, D = DISCO, AFS= Mean of Mechanical Turker AFS scores, our gold standard. For example, NLRD means a combination of NGRAM, LIWC, ROUGE and DISCO.

It is also interesting to examine the differences in model scores for particular argument pairs as shown in Table 4. The best performing model for each row is in **bold** in Table 4. As described in the HIT instructions in Fig. 6, values of AFS near 0 (Row 1) indicate different topics and no similarity. Values near 1 indicate same topic but different arguments (Rows 2,3). Values of 3 and above indicate same FACET (Rows 7,8), and values near 5 are same facet and very similar argument (Rows 12 and 13). Both Arg1 and Arg2 in Row 10 makes the same argument but Arg1 includes additional argumentation. In Row 12, there is very low Ngram overlap, but strong AFS and NLRD performs better than the other models, and LIWC performs well by itself.

In Row 1, UMBC performs the best with a predicted score of 0.37 as opposed to an AFS score of 0.00. Other rows where UMBC on its own provides the best performance are highlighted in the table with Arg1 and Arg2 in **bold**. The top performance of NLRD in Row 5 without UMBC perhaps arises from the semantic information that extermination and holocau are somehow related. NGRAM overlap does the best in Row 13 despite the fact that the phrase *No one argues the point that* does not participate in the NGRAM overlap.

## 4 Related Work

Our approach draws on three different strands of related work: (1) argument mining; (2) semantic textual similarity; and (3) dialog summarization, which we discuss and compare with our work below.

**Argument Mining.** The study of the structure of arguments has a long tradition in logic, rhetoric and psychology (Walton et al., 2008; Reed and Rowe, 2004; Walton, 2009; Gilbert, 1997; Jackson and Jacobs, 1980; Madnani et al., 2012). Much of this work has been on formal (legal or political) argumentation, and the small computational literature that has applied the rhetorical categories of this research has likewise focused on formal, monologic text (Feng and Hirst, 2011; Palau and Moens, 2009; Goudas et al., 2014). More recent work (Ghosh et al., 2014) has attempted to apply these theories to dialogic text in online forums. Ghosh et al. label spans in conversations with attacking moves (CALLOUTS) and their corresponding argumentative TARGETS in another speaker's utterance, and they attempt to learn these callout-target pairs in a supervised framework. Other work attempts to identify general categories of speech-acts such as disagreements or justifications (Misra and Walker, 2015; Biran and Rambow, 2011).

What unites all of the above approaches is an interest in understanding the detailed rheotrical structure of a particular linguistic interaction (monologic or dialogic). Our present work is focused instead on inducing the recurring FACETS in a particular topic domain via weakly supervised learning over several dialogic interactions. Several different threads of recent research on argument mining have strong parallels with this goal (Conrad et al., 2012; Boltuzic and Šnajder, 2014; Hasan and Ng, 2014).

Conrad & Wiebe construct an argument mining system on monologic weblog and news data about universal healthcare. One component of their system identifies ARGUING SEGMENTS and the second component labels the segments with the relevant stance-specific ARGUMENT TAGS. They show that distributional similarity features help identify arguments that belong to the same tag set (notably, we did not find distributional similarity helpful for **AFS**.) Boltuzic & Snajder pursue argument mining on comment streams. Instead of hand-generating argument tags like Conrad & Wiebe, they select short sentential summaries of the key arguments for a given topic from a debate website, and then label comments on the same topic from a different website with the most closely matching summary. The same problem on debate posts is tackled as a "reason classification" problem (Hasan and Ng, 2014), with a probabilistic framework for argument recognition (reason classification) that operates jointly with the related task of stance classification.

All of these approaches differ from ours in three respects. First, they all assume a finite set of topic-specific labels that are determined in some form by the researchers themselves. In contrast, we seek to uncover popular facets via clustering the central propositions across the dialogs. After our own initial categorical efforts, we feel that the argument "topics" have such nuance that they resist clear labels or category membership. Instead, we feel that a scale such as AFS is a better fit, both for the diversity of the data itself and for the idea of inducing FACETS bottom up. Second, these approaches assume the labels are dependent on a particular *stance* towards an issue, whereas our facets are deliberately designed to unify across stance disagreement. Finally, all other approaches in argument mining work from the source text itself. We instead (to our knowledge, for the first time) work from human summaries of dialogs because it is an open question whether the CENTRAL PROPOSITIONS for a dialog are really identifiable as continuous spans of text in the dialog itself. (Indeed, our corpus will allow us to determine how true that assumption is.)

**Semantic Textual Similarity.** There appears to be similarity between FACET induction and aspect learning in sentiment analysis (Brody and Elhadad, 2010), but FACETS are propositional abstract objects, while aspects can usually be described as nouns or properties. Facet induction is more similar to work on **STS** (Mihalcea et al., 2006; Yeh et al., 2009; Agirre et al., 2012; Han et al., 2013; Jurgens et al., 2014). Calculating similarity is a central aspect of AFS. Our scale and MT task for AFS was inspired by the STS task and definition. In addition, as a baseline we apply an off-the-shelf system that calculates STS (UMBC) and compare it with our own system (Han et al., 2013). In order to avoid asking for judgements for many unrelated arguments (CENTRAL PROPOSITIONS), and to make the AFS task more doable for Turkers, we also use UMBC as a filter on pairs of CENTRAL PROPOSITIONS as part of making our HIT. This biases the distribution of the training set to having a much larger set of more similar pairs, which has been a problem for previous work (Boltuzic and Šnajder, 2014), where the vast majority of pairs that were labelled were unrelated. However the AFS task is clearly different than STS, partly because the data is dialogic and partly because it is argumentative. Our results show that we can improve on STS systems for the AFS task.

**Dialog Summarization.** Much previous work on dialog summarization focused on extracting phenomena specific to meetings, such as action items or decisions (Murray et al., 2006; Hsueh and Moore, 2008; Whittaker et al., 2012; Janin et al., 2004; Carletta, 2007). Other approaches, like our work, use semantic similarity metrics to identify the most central or important utterances of a spoken dialog (Gurevych and Strube, 2004), but do not attempt to find the FACETS of a set of arguments across multiple dialogs. Another parallel may exist between work on nuclearity in RST and its use in summarization (Marcu, 1999). However our notion of a CENTRAL PROPOSITION is different than nuclearity in RST, since FACETS are derived from CENTRAL PROPOSITIONS that rise to the top of the pyramid across summarizers, and then (via AFS) across many dialogs on a topic, while RST nuclearity is only defined for a span of text by a single speaker.

Other work examines how social phenomena affect summarization, such as a study of how the politeness level in computer-generated dialogs impacted summaries (Roman et al., 2006). Emotion naturally occurs in the IAC, and summarizers' orientation to emotion is intriguing. Emotional information has been observed even in summaries of professional chats discussing technology (Zhou and Hovy, 2005). However the instructions to our Pyramid annotators were to not include information of this type in the pyramids. We are currently collecting an additional summary corpus using a method that we expect to result in more evaluative and emotional assessments in summaries.

## 5 Conclusion

This paper presents a method and results for extracting FACETS of a topic, across multiple informal arguments on the same topic. We first use human summarization of dialogs as a probe to determine the CENTRAL PROPOSITIONS of each dialog. Then we use clustering in combination with measures of SEMANTIC SIMILARITY to group the CENTRAL PROPOSITIONS into the important FACETS of an argument across many different dialogs. Importantly, we **do not** attempt to enumerate the possible FACETS for an argument in advance, believing that bottom-up discovery of FACETS is a better fit to the problem.

This paper contributes to the current state of knowledge in three ways: (1) we collected summaries of spontaneously-produced written dialog of high social and political importance (available from http://nlds.soe.ucsc.edu/summarycorpus). (2) we proposed a novel application of the pyramid summarization scheme to the task of FACET induction; and (3) we introduce a new task of ARGUMENT FACET SIMILARITY (**AFS**) aimed at identifying FACETS across opinionated dialogs and show that we can identify AFS with a correlation of .54 as opposed to a baseline of .46 provided by a system designed for a similar task. We suspect that the summarize-and-collate approach used here could be promisingly applied to produce annotations on a range of subjective, holistic properties of dialog.

In future work, we aim to expand on this work in several ways. First, we hope to expand summaries, similarity judgments, and systems to several topics beyond gay marriage. We believe, for example, that the features and the system we have trained for AFS will apply to other domains without retraining, since none of the features are topic specific, but we have not shown that. In addition, we aim to develop additional features and improve on the results reported here. For example, we believe that it is possible that other off-the-shelf systems, such as for example one for sentence specificity (Louis and Nenkova, 2011; Louis and Nenkova, 2012), might possibly help with aspects of this task. In addition, in future, we aim to automatically identify CENTRAL PROPOSITIONS without the mediation of human summarizers and evaluators. Given the summaries that we have collected for each dialog, we plan to examine the relationship between the contributors to the related pyramid and the original source text, to determine whether indeed there are surface features of the source that would allow us to treat CENTRAL PROPOSITION detection as an extractive task.

**Acknowledgments** This work was funded by NSF GRANT IIS-1302668, Grant NPS-BAA-03, and an IARPA Grant on Persuasion in Dialogue to UCSC by subcontract from the University of Maryland.